\documentclass[a4paper,11pt]{amsart}
\pdfoutput=1
\usepackage[foot]{amsaddr}
\usepackage{mathtools}
\usepackage{booktabs}

\usepackage{xcolor}
\usepackage[figuresright]{rotating}
\usepackage{url}
\usepackage{algorithmic}
\usepackage{algorithm}
\usepackage{amsmath,amsthm}

\calclayout

\usepackage{booktabs} 
\usepackage{multirow} 

\title{A case study of Consistent Vehicle Routing Problem with Time Windows}

\author{Hern\'an Lespay$^1$}\thanks{H.L. gratefully acknowledges financial support from CONICYT + PAI/CONCURSO NACIONAL TESIS DE DOCTORADO EN EL SECTOR PRODUCTIVO, 2017 + Folio T7817120007.}

\author{Karol Suchan$^{2,3}$}\thanks{K.S. gratefully acknowledges financial support from Programa Regional STICAMSUD 19-STIC-05.}
\address{$^1$Universidad Adolfo Ib\'a\~nez, Santiago, Chile.}
\address{$^2$Universidad Diego Portales, Santiago, Chile}
\address{$^3$AGH University of Science and Technology, Krakow, Poland}
\email{hlespay@alumnos.uai.cl, karol.suchan@mail.udp.cl}

\begin{document}

\begin{abstract}\small
We develop a heuristic for the Consistent Vehicle Routing Problem with Time Windows (ConVRPTW), which is motivated by a real-world application at a food company's distribution center. Besides standard VRPTW restrictions, ConVRPTW assigns each customer just one driver to fulfill their orders during the whole multi-period planning horizon. For each driver and period, a route is sought to serve all their customers with positive demand. For each customer, the number of periods between consecutive orders and the ordered quantities are highly irregular. This causes difficulties in the daily routing, negatively impacting the service level of the company. Similar problems have been studied as ConVRP, where the number of drivers is fixed a priori, and only the total travel time is minimized. Moreover, the clients present no time window constraints, but the visits should be scheduled with a small arrival time variation. In our model, the objective is to minimize the number of drivers. We impose hard time windows but do not consider time consistency in more detail. We compare solutions given by the heuristic with solutions of a MILP model on a set of small artificial instances and solutions used by the food company on real-world instances. The results show the effectiveness of the heuristic. For the company, we obtain significant improvements in the routing plans, with a lower number of vehicles and a higher rate of orders delivered within the prescribed time window.
\end{abstract}



\maketitle

\section{Introduction}
The problem studied in this paper is motivated by a real-world application at a food distribution company in Chile. This company is one of the leading national producers and marketers of meat and poultry. The company has seventeen distribution centers throughout the national territory. In this paper, we focus on one of their distribution centers. However, this study can be replicated in each distribution center of the company because their operations are quite similar throughout the country. The company has large sales volumes nationwide, and the costs associated with distribution activities represent an essential component of their total expenditures. Therefore, a reduction of 5 or 10 percent in the routing costs would mean a significant improvement for the company.

The customers of the company correspond to supermarkets, restaurants, wholesalers, and small convenience stores spread throughout a region of Chile. The set of customers changes, on average, by 10\% from one month to another. This customers' variability is in large part due to the low frequency and irregularity of orders from small convenience stores, which represent 60\% of the total number of customers. Also, the customers are characterized by high variability of service times and, even under perfect conditions, the time needed to unload the delivery can differ by about an hour from one customer to another. Additionally, each customer requires that their orders be delivered within a certain time-window.

The company has a fleet of vehicles, which are dispatched from the distribution center to their respective customers. Each vehicle has a fixed subset of customers assigned to it, and each day the vehicles serve their corresponding customers.

Due to the high variability in the activity of the customers, the quantities they order and the service times, the company currently experiences significant problems in the fulfillment process during the periods of high activity: the customers time windows may not be respected and, in some cases, the delivery even has to be postponed to another day. Consequently, it is vital for the company to develop a decision support system that would improve the planning and scheduling of the distribution of products to its customers, seeking the right balance between operational costs and service quality.

\subsection{Consistency in Vehicle Routing}
In the standard version of the vehicle routing problem (VRP), we have a set of customers with a known demand and a set of homogeneous vehicles with a limited capacity based at a depot. The goal is to determine a set of routes to fulfill all orders at minimum cost. The VRP was introduced in \cite{dantzig1959}, and the objective is oriented to increase the profitability of the companies without considering customer satisfaction, an important concern in competitive markets.

Solving the vehicle routing problem for each day in an independent way allows companies to minimize operational costs and efficiently use their vehicle fleet. However, this has some operational disadvantages. Frequent changes in the vehicle's routes can negatively impact the productivity and job satisfaction of the drivers. Consequently, due to the growing industry competition, it is crucial to balance the objective of designing routes of low cost with that of providing high-quality service to increase customer satisfaction.

In general, customer satisfaction is directly related to providing a consistent service (see \cite{kovacs2014b}). Service consistency has three different dimensions: i) Time, it refers always to visit a given customer at roughly the same time, ii) Person, it refers always to execute the delivery by the same driver, and iii) Delivery, it refers to minimizing the variability of the quantity delivered when the supplier decides the quantity delivered (vendor-managed inventory systems). The relevance of each one of these three dimensions of consistency changes from one market to another.

From the point of view of distribution companies, consistent routes improve driver satisfaction and productivity, because they allow drivers to gain knowledge about the territory where they are working, the customers' formal receiving procedures and more informal preferences, and get personally acquainted with the employees in charge of coordinating the deliveries on the customers' sites. All this improves drivers' abilities to overcome difficulties resulting from unexpected incidents in transit and on customers' sites.

Companies have to choose between routing plans of minimum cost, which are not consistent, and routing plans of the higher cost that consider consistency. In \cite{wong2008}, it is argued that a higher total distance in the design of the routes to privilege the service consistency is more than compensated by a greater satisfaction of customers and drivers, and a productivity increase of the drivers. Consequently, we aim to find a set of consistent routes with the lowest operational cost.

\subsection{Problem Characteristics}
In the literature, this kind of situation is often modeled as the Consistent Vehicle Routing Problem (ConVRP). ConVRP was introduced in \cite{groer2009}, motivated by the importance of providing high-quality customer service in the small package shipping industry. It takes into account time and person consistencies.

The small package shipping industry operates in conditions that differ from the ones present at the company under study in several aspects.
First, companies in the small package industry admit that it is important for recurring customers to receive service at about the same time over different days. This is not the case for the company under consideration. Here, the customers appreciate that the deliveries arrive within their time windows, independently of the precise moment within this time-range that the delivery occurs.

Second, small package shipping companies have short and relatively constant service times in most of the cases. However, many customers of the company under study need longer service times and present a considerable risk of extending them, even more, when an operational incident occurs.

Third, small package shipping companies prefer to have the same service provider for a group of customers to generate familiarity with a single geographical zone and use it to improve travel times. In the case under study, the company wants to generate familiarity between the driver and the person who receives the deliveries in order to reduce the risk of long service time or denial of reception due to some incident.

Finally, the characteristics of the fleet are different. It is common in small package shipping companies that the vehicle fleet is proprietary, the vehicles are small, and there is only one person operating the vehicle. Instead, at the company under study, the vehicle fleet is outsourced from many transportation companies, the vehicles are large (capacity of 5 tons.), and they are operated by a team of one driver and two assistants. Moreover, the vehicles must have a refrigeration system because of the characteristics of the products delivered. Therefore, it is important for the company to execute the deliveries utilizing a reduced number of vehicles to minimize the operational costs.

\subsection{Modeling Consistency}
One of the most general variants of ConVRP studied in the literature was proposed in \cite{kovacs2015b}, which is based on \cite{kovacs2015a}. The results of this reference paper offer some flexibility to different operational contexts. However, it is still necessary to make some adjustments to solve the company's real problem under study. Next, we highlight the main differences.

On the one hand, motivated by the small package delivery, the objectives of the reference paper is to optimize the routing cost, arrival time consistency, and driver consistency in a multi-objective approach. For our purposes, these objectives are not adequate. Our principal concern is to minimize the number of vehicles. On the other hand, the reference paper uses AM/PM time windows motivated by the requirements of companies that hire part-time workers to handle package delivery. Also, the service times considered are the same for all the customers. In contrast, we need to respect the time windows imposed by a variety of customers with service times that are quite dissimilar. Finally, the reference paper minimizes the number of drivers that serve customers, allowing a customer to be served by more than one driver. However, this approach is not convenient for our purposes. Many customers of the company have reception areas where many vehicles arrive simultaneously to deliver different products. The queue management tends to be somewhat informal and, if there is some familiarity between the driver and the person coordinating the reception area, the waiting times tend to be shorter, and the general efficiency of the process improves.

Consistency requirements link different days in the planning horizon, making it impossible to decompose the multi-period problem into several independent one-period problems. In the literature, most approaches to solve ConVRP problems focus on heuristic methods, because exact methods of mixed-integer linear programming allow only to solve very small instances. For example, \cite{groer2009} solved instances up to 12 customers over three days, and \cite{goeke2019} solved instances with 30 customers over five days. The exact methods are computationally intractable for the instances of the company under study since we need to solve instances with about two thousand customers over a planning horizon of at least 25 days.

\subsection{Our Contribution}

The contributions of this paper are four-fold. Firstly, we model a real-life case of a distribution problem at a major Chilean food company as ConVRPTW, a variant of ConVRP that incorporates time windows of the customers and tries to minimize the number of vehicles used to form consistent routing plans, under high variability in client activity and quantities ordered. To the best of our knowledge, there are very few results in the ConVRP-related literature that consider the restriction of time windows and the objective of minimizing the number of vehicles. Moreover, they are limited to instances of much smaller size and lesser variability of the demand. The most common setting is that all clients are active in all periods of the planning horizon, and only their demand varies according to some continuous distribution - in our case, most clients are active only in a few periods, that moreover are scattered randomly over the planning horizon (not respecting any known distribution). Secondly, we propose an efficient heuristic (inspired by \cite{nagata2009}) for reducing the number of routes in ConVRPTW. This heuristic is based on the holding list strategy (see \cite{lim2007}), and uses a guided local search to guide the ejections. Thirdly, we propose a simple and fast heuristic to find a feasible initial solution. Finally, we provide some managerial insights by developing an analysis of assigning customers-to-drivers using historical data in order to evaluate the proposed solution of ConVRPTW in a real-life context. We analyze the quality of the client-to-driver (tactical) assignments generated by our algorithm run on historical data as the basis for the operational day to day routing under the conditions of uncontrolled variability of demand experienced by the company under study.

The remainder of this paper is organized as follows. Section 2 introduces the literature review of ConVRP.  Section 3 presents a description of the problem. Section 4 describes the solution framework. Section 5 reports the results of computational experiments. Finally, section 6 presents the conclusions.

\section{Literature Review}
Generating template routes is a common strategy to solve ConVRP. It was used in \cite{groer2009,tarantilis2012,kovacs2014a,xu2018}. To satisfy the consistency requirements, this strategy uses the precedence principle: \textit{if a driver services $a$ before $b$ on one day, then $a$ has to receive service before $b$ from the same vehicle on all days that they both require service}. Intuitively, routes that follow this precedence principle should tend to comply not only with the person but also the time consistency. The template strategy consists of two stages. In the first stage, the problem is reduced to construct routes for a subset of customers, called frequent customers. A customer that requires service on more than one period during the planning horizon is called {\em frequent}, and {\em non-frequent} otherwise. Then, the mean demand over the planning horizon is calculated for each frequent customer and used to create a unique VRP instance. The set of resulting routes is called the template. Each route of the template is assigned, one-to-one, to a driver. In the second stage, daily routes are generated by replicating the template for each day, with the original values of customer demands on a particular day. Finally, the daily routes are updated by deleting frequent customers with zero demand and inserting non-frequent customers with positive demand.

The template approach restricts the search space, and therefore, good solutions can be found quickly. Yet, better solutions might be missed (see \cite{kovacs2015a}), and even more, the optimal solution may be unreachable. Instead of working with template routes, we design a solution strategy that operates on the entire routing plan, an approach which also is used in
\cite{kovacs2015a,kovacs2015b,lian2016,braekers2016,stavropoulou2019,campelo2019}.

In the following subsections, we describe the main works in the literature to solve ConVRP and its variants, using these strategies. For a review of works on VRPTW, we refer the reader to one of many surveys on the topic present in the literature (for example, see \cite{gendreau2010}).

\subsection{Template Based ConVRP}
In \cite{groer2009}, the authors proposed a variant of the traditional VRP that considers several routing periods with fixed customer demand assigned to each of them, and with consistency considerations in two dimensions: time and person. Thus, they defined ConVRP as the problem of finding a set of minimum cost routes to service a set of customers with known demands over multiple periods considering the classical vehicle routing constraints complemented with time and person consistency requirements, i.e., each customer has a fixed driver assigned, that delivers the orders at roughly the same time on each day that the customer needs service. Therefore, the routes that visit a given customer over different periods have to be all assigned to the same driver. Both time and person consistencies are treated as hard constraints: the maximum arrival time difference of each customer is bounded, and only one driver per customer is allowed. They generated template routes using a multi-start algorithm based on the record-to-record approach for solving the problem.

In \cite{tarantilis2012}, a Tabu Search (TS) algorithm was proposed to minimize the total travel time, modifying both the template routes and the actual daily schedules sequentially. In \cite{kovacs2014a}, an Adaptative Large Neighborhood Search (ALNS) algorithm was used to solve the problem. Additionally, a relaxed variant of ConVRP was introduced, in which the departure times from the depot can be delayed to adjust the arrival times at the customers' sites. The authors demonstrated that this slight relaxation of the starting times leads to improved time consistency, while the travel times remain almost unchanged. In \cite{xu2018}, a Variable Neighborhood Search (VNS) algorithm was presented. The proposed algorithm consists of two stages. In the first stage, a VNS is applied to obtain optimized template solutions. The solutions obtained might be infeasible, but the second stage is applied to make it feasible and improve it further.

\subsection{Non-template Based ConVRP and Related Problems}

In \cite{goeke2019} an exact method for ConVRP based on the column generation technique was proposed. This method was used for solving medium-sized instances with five days and 30 customers. As an upper bounding procedure, a large neighborhood search (LNS) was developed, featuring a repair procedure specifically designed to improve the arrival-time consistency of solutions. Moreover, the LNS used as a stand-alone heuristic can improve the solution quality on benchmark instances from the literature. In \cite{rodriguez2019} an integer linear programming formulation and several families of valid inequalities for the Periodic Vehicle Routing Problem with Driver Consistency (PVRP-DC) were studied. The problem was solved using an exact branch-and-cut algorithm, and computational results were shown on a wide range of randomly generated instances with a maximum size of five days and 71 customers. In \cite{luo2015} a mixed integer programming model for the multi-period vehicle routing problem with time windows and limited visiting quota (MVRPTW-LVQ) was proposed, which requires that any customer can be served by at most a certain number of different vehicles over the planning horizon. A three-stage approach to solve the MVRPTW-LVQ was proposed. Stage one finds initial solutions using the decomposition algorithm. Stage two reduces the number of vehicles in the solution by employing a repair procedure in a tree search algorithm. Stage three performs a tabu search post-optimization procedure that focuses on reducing the total travel distance. Computational experiments on benchmark data showed improvements on different data sets. In \cite{braekers2016}, person consistency in a multi-period dial-a-ride problem (DC-DARP) was explored. Person consistency is considered by bounding the maximum number of different drivers that transport a user over a multi-period planning horizon. The authors proposed different formulations and assessed their efficiency using a branch-and-cut scheme. Additionally, they proposed an ALNS that generates near-optimal solutions.

In \cite{kovacs2015a}, it was pointed out that the consistency requirements of ConVRP may be too restrictive. Thus, a relaxed ConVRP, called Generalized ConVRP (GenConVRP), was introduced. Here, a customer may be served by a limited number of drivers instead of only one. The time consistency is not enforced by constraints but is penalized in the objective function. Moreover, the precedence principle is not enforced, so instead of assigning some kind of ``a priori'' routes to the drivers, they only get fixed customer sets to service (with no precedence constraints) - each day is routed (sequenced) separately, but preserving the time consistency between days. Vehicle departure times may be adjusted to obtain stable arrival times. Apart from the relaxations mentioned above, customers are associated with AM/PM time windows motivated by the requirements of companies that hire part-time drivers to handle package delivery. The authors proposed a Large Neighborhood Search algorithm (LNS) to solve GenConVRP.

In \cite{kovacs2015b}, a multi-objective optimization approach was proposed. GenConVRP is extended by considering different objective functions. Routing cost, time consistency, and person consistency are considered as independent objectives of the problem. Two exact approaches based on the $\epsilon$-constraint framework were proposed for small test instances. For large instances, the authors proposed a Multi-Directional Large Neighborhood Search (MDLNS) that combines the Multi-Directional Local Search framework and the LNS for the problem. Another multi-objective variant of ConVRP was presented in \cite{lian2016}. Instead of modeling consistency requirements as hard constraints, they are included as objectives. An improved version of the multi-directional local search (MDLS) was developed to solve the problem.

In \cite{feillet2014}, in contrast to the classical ConVRP, where the maximum arrival time difference of each customer is bounded, the authors proposed to minimize the maximum number of time classes in which a customer is visited. For visits grouped into a single time class, the difference between the earliest and the latest visit is not larger than a given bound. Person consistency is not considered in the proposed solution approach, but the number of drivers per customer is reduced in a post-processing step.
Also, in \cite{hernandez2017}, it is highlighted that offering more and tighter time slots is perceived by customers as improved service quality, but care should be taken about the impact it may have on the transportation costs. The topic was studied in the context of a tactical problem, where a time slot schedule for delivery service over a given planning horizon must be selected in each zone of a geographical area.

In \cite{stavropoulou2019}, ConVRP with Profits was introduced. There are two sets of customers, the frequent customers that are mandatory to service, and the non-frequent potential customers that are optional, with known and estimated profits, respectively, both having known demands and service requirements over a planning horizon of multiple days. The objective is to determine vehicle routes that maximize the net profit while satisfying vehicle capacity, route duration, and consistency constraints. An Adaptive Tabu Search was developed to address this problem, utilizing both short- and long-term memory structures to guide the search process. In \cite{campelo2019}, ConVRP with different service level agreements was presented. In this variant of ConVRP, customers have several shipping periods during the day, time windows, and different release dates. A metaheuristic algorithm was developed to solve this problem on larger instances. In \cite{zhen2019} a new variant that considers simultaneous distribution and collection (ConVRPSDC) was introduced, for which a mixed-integer programming model was formulated. To solve the problem, three heuristics were proposed based on the record-to-record (RTR) travel algorithm, the local search with variable neighborhood search (LSVNS), and the tabu search-based method. The results showed that the RTR-based heuristic has an advantage in small-scale instances. However, for medium-scale instances, the best option is LSVNS-based heuristic, which can solve instances with 40 customers and 5 days within 10 seconds. Moreover, LSVNS-based heuristic can solve large-scale instances with 200 customers and 5 days within 3 hours.

\section{The Problem}
The ConVRPTW problem is defined on a complete directed graph $G=(N,A)$, where $N=\{0,1,...,n\}$. $0$ represents the depot, and positive integers represent the customers. $A=\{(i,j)| i,j \in N, i\neq j\}$ is the set of arcs. Each arc $(i,j) \in A$ is associated with a distance $\delta_{ij}$ and a travel time $t_{ij}$, with  $\delta_{ij} \neq t_{ij}$, for all $(i,j) \in A$. We assume that distances and travel times satisfy the triangle inequality. Customers are visited on routes traversed by a homogeneous fleet of vehicles in the set $K$. The number of vehicles is not restrictive (i.e., $|K| = |N \setminus \{0\}|$).

Each vehicle $k \in K$, with given capacity $Q$ is located at the depot, from where it departs at time $0$, and where it must return before time $T$. The planning horizon involves $|D|$ days, where $D$ is the set of days. On each day $d \in D$, each customer $i \in N \setminus \{0\}$ has a demand $q_{i,d}$ and service time $s_i$. We assume that $q_{i,d} \leq Q$ and it is not possible to perform split deliveries (because of the person consistency requirement). We use auxiliary parameters $w_{i,d}$ equal to $1$ if customer $i$ requires service on day $d$ ($q_{i,d} > 0$), and equal $0$ otherwise.

Person consistency is enforced by assigning each customer $i \in N \setminus \{0\}$ to the same vehicle $k$ over the entire planning horizon. Time consistency is limited only to respecting the time windows: the service start-time can vary freely within the interval $[l_i,u_i]$ defined as the customer time window, where $u_i$ and $l_i$ are the upper and lower bound of the time window of the customer $i$, $i \in N \setminus \{0\}$. We only require that the service start-time has to fall within the time window, not the sum of service start-time and service times - a vehicle is allowed to complete the delivery after the time window closes. For the ease of presentation, we identify each vehicle $k$ with the set of customers assigned to it (we will speak of a vehicle and the set of customers assigned to it interchangeably).

A solution $R$ of a ConVRPTW instance is determined by $|K|$ subsets of customers: each one of them, $k\in K$, with an independent route (the order of visits for frequent customers is not preserved from one day to another), $r_{k,d}$, for every day, $d\in D$, when some of the customers in $k$ require service. The objective is to minimize the number of non-empty sets of customers, such that i) the total quantity carried by each vehicle on each day does not exceed the capacity $Q$, ii) the time windows of all customers and the depot are respected, and iii) each customer is served by exactly one driver over the planning horizon. The customers are not visited on the days when their demand is $0$. Empty subsets of customers are not assigned to drivers, so the number of vehicles used over the planning horizon is decided in the model.

We present a MIP formulation for ConVRPTW in the following model. The model uses the binary decision variables $x_{i,j,k,d}$ that equal $1$ if arc $(i,j)$ is traversed by vehicle $k$ on day $d$, and $0$ otherwise. The binary decision variables $y_{i,k,d}$ equal $1$ if customer $i$ is assigned to vehicle $k$ on day $d$, and $0$ otherwise. The binary decision variables $z_{k}$ equal $1$ if vehicle $k$ is used in any day of the planning horizon, and $0$ otherwise. Finally, the continuous decision variables $a_{i,d}$ equal the service start-time at customer $i$ on day $d$, and $0$ if no service for customer $i$ is required on day $d$. We define the service start-time at customer $j$ on day $d$ as $a_{jd} = a_{id} + s_i + t_{ij} + e_{jd}$, where $e_{jd}$ is equal to the waiting time if the vehicle arrives at customer $j$ on day $d$ earlier than $l_j$, and $0$ otherwise.

\begin{align}
&Minimize \quad \sum_{k \in K} z_k
\end{align}
subject to
\begin{align}
&\sum_{k \in K} y_{i,k,d} = w_{i,d}    & \forall i\in N \setminus\{0\}, d \in D\\
&\sum_{i\in N \setminus\{0\}} q_{i,d} y_{i,k,d} \leq Q    &\forall k\in K, d \in D\\
& z_k \geq y_{i,k,d}  	& \forall i\in N \setminus\{0\}, k \in K, d \in D\\
&\sum_{i \in N\setminus \{j\}} x_{i,j,k,d} = \sum_{i \in N\setminus \{j\}} x_{j,i,k,d} = y_{j,k,d}  &\forall j\in N \setminus \{0\}, k\in K, d \in D\\
&\sum_{j \in N \setminus \{0\}} x_{0,j,k,d}= \sum_{i \in N \setminus \{0\}} x_{i,0,k,d} = z_k  & \forall k \in K, d\in D\\
&w_{i,\alpha} + w_{i,\beta} - 2 \leq y_{i,k,\alpha} - y_{i,k,\beta}  & \forall i\in N \setminus\{0\}, k\in K, \alpha,\beta \in D, \alpha \neq \beta\\
&a_{i,d} + x_{i,j,k,d}(s_i + t_{i,j} + e_{jd}) - (1-x_{i,j,k,d})T \leq a_{j,d}  &\forall i\in N, j\in N \setminus\{0\}, i \neq j, k\in K, d \in D\\
&a_{i,d} + x_{i,j,k,d}(s_i + t_{i,j} + e_{jd}) + (1-x_{i,j,k,d})T \geq a_{j,d}  &\forall i\in N, j\in N \setminus\{0\},i \neq j, k\in K, d \in D\\
&a_{i,d} + s_i + t_{i,0} \leq T  & \forall i\in N \setminus\{0\}, d \in D\\
&l_i w_{i,d}\leq a_{i,d} \leq u_i w_{i,d} & \forall i \in N, d \in D\\
&z_k \in \{0,1\}  &\forall k \in K\\
&x_{i,j,k,d} \in \{0,1\}   & \forall i,j \in N,i \neq j, k\in K, d \in D\\
&y_{i,k,d} \in \{0,1\}   & \forall i \in N, k\in K, d \in D\\
&a_{i,d} \geq 0   & \forall i \in N, d \in D
\end{align}

The objective function $(1)$ minimizes the number of vehicles. Constraints $(2)$ guarantee that each customer is serviced on each day they require service, and inequalities $(3)$ make sure that the vehicle capacity is not exceeded. Constraints $(4)$ ensure that if customer $i$ is served by vehicle $k$ on the day $d$, then vehicle $k$ must be used during the planning horizon. Constraints $(5)$ ensure that all assigned customers have exactly one predecessor and one successor, and equalities $(6)$ are flow conservation constraints on the depot for each vehicle $k$ and day $d$.

Person consistency is guaranteed in $(7)$. So, if customer $i$ requires service on days $\alpha$ and $\beta$, then $w_{i,\alpha} + w_{i,\beta} - 2 = 0$, enforcing person consistency by $y_{i,k,\alpha}=y_{i,k,\beta}$. Inequalities $(8)$ and $(9)$ set the service start-times at the customers. So, if customer $j$ is visited after customer $i$ for some vehicle $k$ and day $d$, then $x_{i,j,k,d} = 1$ and we obtain $a_{i,d} + s_i + t_{i,j} + e_j = a_{j,d}$ from constraints $(8)$ and $(9)$. Otherwise, the arrival times for two customers are not related.


Constraints $(10)$ enforce that vehicles return to the depot on time. The time windows of the customers are enforced by inequalities $(11)$. Finally, the integrality and non-negativity requirements of the decision variables are guaranteed by constraints $(12)-(15)$.

\section{Solution Framework}
\begin{figure}[th!]
  \centerline{\includegraphics[scale=0.7]{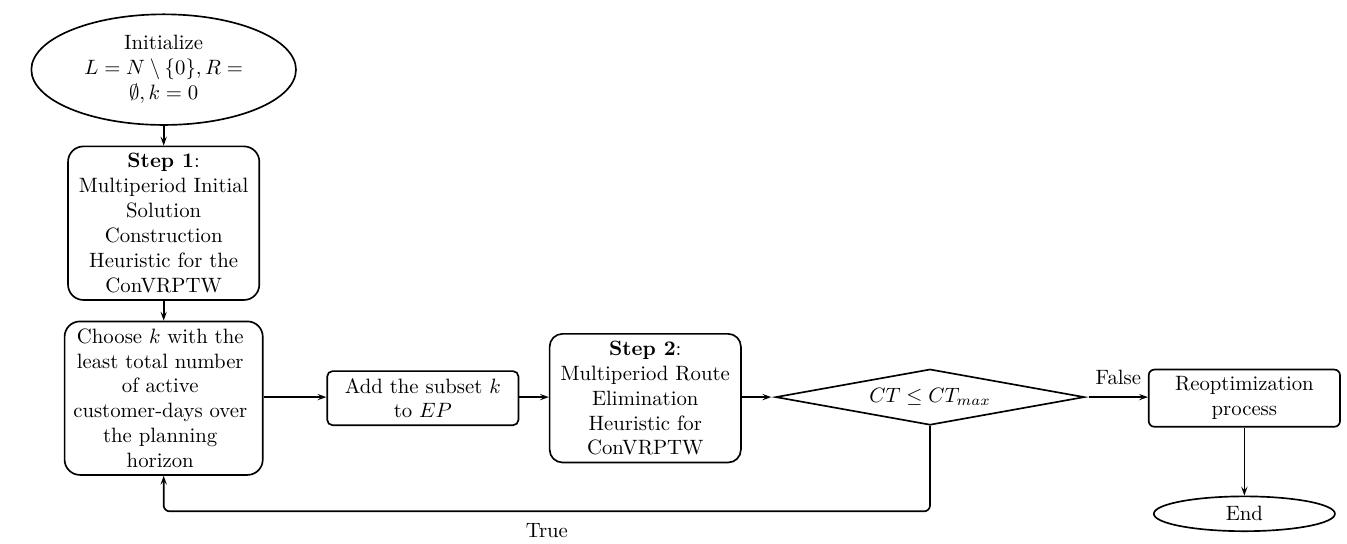}}
\caption{Heuristic for ConVRPTW\label{fig:1} }
\end{figure}

We propose an efficient heuristic for minimizing the number of routes in ConVRPTW. The heuristic is composed of two different steps. In the first step, based on the insertion heuristic proposed in \cite{solomon1987}, we construct a feasible initial solution for ConVRPTW. In the second step, inspired by \cite{nagata2009}, we develop a heuristic focused on reducing the number of routes in the solution obtained in the first step.
Finally, we reoptimize the final solution obtained in step two with respect to the total travel distance. This reoptimization process consists of sequentially applying the well-known intra-route 2-opt and inter-route relocation (adapted to the multi-period case) local operators until a local optimum is reached (see \cite{braysy2005a}).
We describe the two steps in detail in the following subsections. See Figure \ref{fig:1} for a general overview.

\subsection{Initial Solution Construction}
The algorithm we use to create an initial solution for ConVRPTW is an extension of the well-know one-period insertion heuristic proposed in \cite{solomon1987} to work with multiple periods.

The one-period heuristic is a sequential insertion heuristic for constructing solutions to the vehicle routing problem with time windows (VRPTW). It has low time complexity and tends to generate solutions with a small number of vehicles, compared to other simple construction heuristics (see \cite{braysy2005a}). Because the main purpose of the first step of our algorithm is to construct a feasible initial solution for ConVRPTW quickly, a multi-period extension of the heuristic mentioned above seems a good option. To the best of our knowledge, no previous works in the area of ConVRP used this approach.

In the following, we describe both the one-period initial solution construction heuristic for the VRPTW proposed in \cite{solomon1987} and our multi-period extension of this heuristic for ConVRPTW.

\subsubsection{One-period Initial Solution Construction Heuristic for the VRPTW}
The heuristic starts with a set $N\setminus \{0\}$ of customers, each of them with their respective time window requirements $[l_i,u_i]$ and service time $s_i$, $\forall i\in N \setminus \{0\}$ ($s_0 = 0$). Each vehicle has capacity $Q$, must leave the depot at time $0$, and return before $T$. Finally, from the distance and travel time matrices, we have the distance $\delta_{i,j}$ and the travel time $t_{i,j}$ between the customers $i$ and $j$, respectively. The output of the heuristic is a feasible initial solution $\sigma$ that contains a set $K$ of routes for the VRPTW.

Firstly, it is necessary to define a criterion for initializing every new route $r$. Two criteria are tested. The first criterion ($ic = 1$) chooses the unrouted customer that is farthest from the depot. This criterion assumes that a customer with a larger distance from the depot is more difficult to insert into an already established route, so they are given priority to start a new route. The second criterion  ($ic = 2$) chooses the unrouted customer with the earliest time window closing time. Complementary to the first criterion, this one assumes that a customer with a short time window closing date is more difficult to insert into an established route, so they are also given priority to start a new route.

Secondly, after initializing the current route $r$, considering that some customers are already assigned to routes in the current partial solution $\sigma$, the heuristic uses two criteria $\hat{c}_1(\sigma,r,v)$ and $\hat{c}_2(\sigma,r)$, where $\hat{c}_2(\sigma,r)$ is a function of $\hat{c}_1(\sigma,r,v)$, to insert a unrouted customer $v$ into $r$ between two adjacent customers. Let $r=(i_1,i_2,...,i_m)$, where $i_1=i_m=0$. For each unrouted customer $v$, we first compute its best feasible insertion place with respect to vehicle capacity and time windows as:
\begin{align*}
&\hat{c}_1(\sigma,r,v) = argmin\{ c_1(\sigma,r,v,p): p=1,...,m-1\},
\end{align*}
where:
\begin{align*}
&c_1(\sigma,r,v,p) = \alpha_1 c_{1,1}(\sigma,r,v,p) + \alpha_2c_{1,2}(\sigma,r,v,p), \alpha_1 + \alpha_2 = 1 ,\alpha_1, \alpha_2 \geq 0,\\
&c_{1,1}(\sigma,r,v,p) = \delta_{i_p,v} + \delta_{v,i_{p+1}} - \mu \delta_{i_p,i_{p+1}}, \mu \geq 0,\\
&c_{1,2}(\sigma,r,v,p) = a^v_{i_{p+1}} - a_{i_{p+1}}.
\end{align*}
with $\alpha_1$, $\alpha_2$ and $\mu$ parameters to be adjusted.

$a^v_{i_{p+1}}$ is the service start-time at $i_{p+1}$ after inserting $v$ just before $i_{p+1}$ on the route, and $a_{i_{p+1}}$ the service start-time at $i_{p+1}$ without inserting $v$.

Finally, the best customer $v$ to be inserted in the route is selected using the second criterion defined as:
\begin{align*}
&\hat{c}_2(\sigma,r) = argmin\{c_2(\sigma,r,v): \text{$v$ unrouted and feasible}\}.
\end{align*}
where:
\begin{align*}
&c_2(\sigma,r,v) = c_1(\sigma,r,v,\hat{c}_1(\sigma,r,v)) - \lambda \delta_{0,v}, \lambda \geq 0.
\end{align*}
with $\lambda$ a parameter to be adjusted.

Customer $v^*= \hat{c}_2(\sigma,r)$ is then inserted in the route between the adjacent customers $i$ and $j$, where $i = i_{\hat{c}_1(\sigma,r,v^*))}$ and $j=i_{\hat{c}_1(\sigma,r,v^*)+1}$. When no more customers with feasible insertions can be found, a new route is started, until all customers have been routed. Once a new route is started, the previously constructed routes are not modified anymore.

This insertion heuristic tries to maximize the benefit derived from servicing a customer on the partial route being constructed rather than on a direct route. The best feasible insertion place for an unrouted customer is the one that minimizes the weighted combination of distance and time insertion costs.

\subsubsection{Multiperiod Extension of the Initial Solution Construction Heuristic for ConVRPTW}

\begin{figure}[th!]
  \centerline{\includegraphics[scale=0.7]{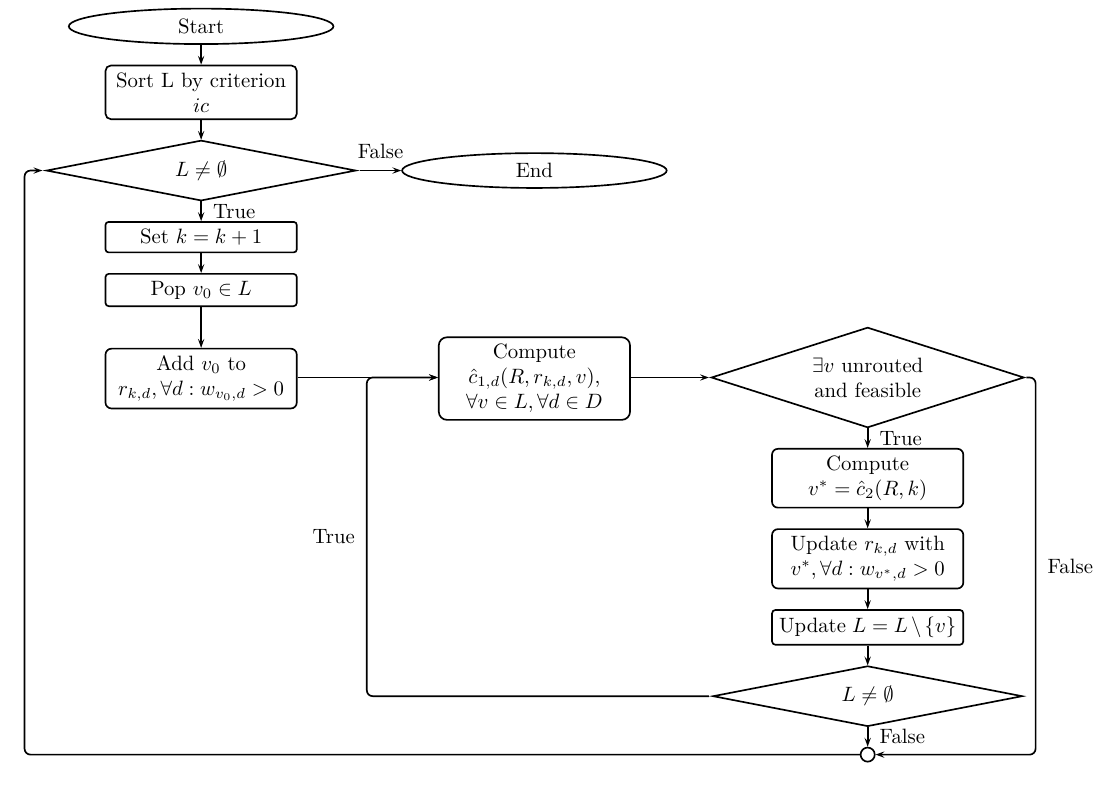}}
\caption{Multiperiod Extension of the Initial Solution Construction Heuristic for ConVRPTW\label{fig:2} }
\end{figure}

Since one-period initial solution construction heuristic for the VRPTW described above can not be used to find an initial solution for ConVRPTW, we propose a multi-period extension for ConVRPTW. See Figure \ref{fig:2} for a general overview.

In section 3, we described that a solution $R$ of a ConVRPTW instance is determined by $|K|$ subsets of customers, and each non-empty subset $k$ of customers is assigned to a vehicle. For each day $d$ of the planning horizon, $d\in D$, and vehicle $k$, $k\in K$, a feasible route $r_{k,d}$ has to be created that services all the customers of $k$ that are active on $d$. In other words, instead of creating groups of customers with just a single feasible route, as the one-period heuristic does, we must find subsets of customers for which a feasible route exists for each day of the planning horizon, serving all the respective {\em active} customers.

In this new context, a feasible (for vehicle capacity and time windows) insertion of a client $v$ into a ``vehicle'' (a subset of customers) $k$ corresponds to a set of feasible insertions of $v$ into routes $r_{k,d}$, for each day $d$ when $v$ has positive demand. The best insertion place is computed in a way similar to that of the one-period heuristic: using $\hat{c_1}(\sigma,r,v)$, but extended for the multi-period case:
\begin{align*}
&\hat{c}_{1,d}(R,r_{k,d},v) = argmin\{ c_{1,d}(R,r_{k,d},v,p): p=1,...,m-1\},
\end{align*}
where:
\begin{align*}
&c_{1,d}(R,r_{k,d},v,p) = \alpha_1 c_{1,1,d}(R,r_{k,d},v,p) + \alpha_2c_{1,2,d}(R,r_{k,d},v,p), \alpha_1 + \alpha_2 = 1 ,\alpha_1, \alpha_2 \geq 0,\\
&c_{1,1,d}(R,r_{k,d},v,p) = \delta_{i_p,v} + \delta_{v,i_{p+1}} - \mu \delta_{i_p,i_{p+1}}, \mu \geq 0,\\
&c_{1,2,d}(R,r_{k,d},v,p) = a^v_{i_{p+1}} - a_{i_{p+1}}.
\end{align*}

Note that the value of these parameters changes depending on the previously computed route for each day of the planning horizon.

Finally, in contrast to the one-period heuristic, we implicitly consider the service frequency of each customer to compute the best customer $v$ to be inserted in a subset $k$ of customers. In order to accomplish this, we redefine $\hat{c_2}(\sigma,r)$ as:
\begin{align*}
&\hat{c}_2(R,k) = argmin\{c_2(R,k,v): \text{$v$ unrouted and feasible}\}.
\end{align*}
where:
\begin{align*}
&c_2(R,k,v) = \frac{\sum_{\{d \in D:q_{v,d > 0}\}} \left[  c_{1,d}(R,r_{k,d},v,\hat{c}_{1,d}(R,r_{k,d},v)) - \lambda \delta_{0,v} \right]}{|\{d \in D:q_{v,d > 0}\}|}, \lambda \geq 0.
\end{align*}

\subsection{Route Elimination Heuristic}\label{subs:reh}
A common characteristic of most approaches for one-period VRPTW problems is to use a route elimination procedure as an independent heuristic component (see \cite{gendreau2010}). A very powerful route elimination heuristic that outperforms other existing approaches was proposed in \cite{nagata2009}. Comparison with other algorithmic approaches shows that this heuristic is one of the most effective and efficient ones, and can obtain, for most groups of problem instances considered in \cite{gendreau2010}, the lowest known cumulative number of vehicles in low total computation time. On the other hand, simplicity and flexibility are both relevant criteria for real-life applications. Simplicity refers to the ease of implementation related to a low number of parameters that need tuning. Flexibility is related to the ability of an algorithm to accommodate various side constraints. The route elimination heuristic proposed in \cite{nagata2009} scores high on both. So, we propose to extend it to a multi-period version for ConVRPTW.

In what follows, we first describe the one-period route elimination heuristic for the VRPTW proposed in \cite{nagata2009}, and then we explain the way to adapt it for ConVRPTW.

\subsubsection{One-period Route Elimination Heuristic for VRPTW}
The heuristic starts with a given feasible initial solution $\sigma$ that contains a set of $|K|$ routes for the VRPTW. This initial solution can be obtained from the one-period initial solution construction heuristic for the VRPTW described above. We repeatedly apply the one-period route elimination heuristic to the incumbent solution $ \sigma $ to reduce the number of routes one by one. Finally, we obtain a new solution $\sigma$ with a lower number of vehicles.

Initially, a route from the current solution $\sigma$ is randomly selected. Its customers are temporarily removed from the solution and transferred to a stack called {\em ejection pool} ($EP$). Penalty counters $p[v], \forall v \in N\setminus \{0\}$, are initialized with value 1. The penalty counter $p[v]$ denotes how many times an attempt to insert $v$ in a route has failed so far. Intuitively, the value of $p[v]$ gives an estimate of how difficult it is to re-insert a customer $v$ into $\sigma$. Then, repeated attempts are made to insert the customers from $EP$ into $\sigma$ until $EP$ is empty or the total computation time $CT$ reaches a given limit $CT_{max}$. This procedure is executed in three stages.

In the first stage, we compute $\mathcal{N}_{insert}(v_{in},\sigma)$, the set of all partial solutions that can be obtained by inserting $v_{in}$ at all insertion positions of different routes in $\sigma$. If $\mathcal{N}_{insert}(v_{in},\sigma)$ contains feasible solutions, then one of them is chosen at random. The current partial solution is updated before proceeding to select a new customer from $EP$. Otherwise, we move forward to the next stage.

In the second stage, an infeasible insertion is chosen from $\mathcal{N}_{insert}(v_{in},\sigma)$ and temporarily accepted, such that $F_p(\sigma) = P_c(\sigma) + \alpha P_{tw}(\sigma)$ is minimum. $F_p$ is a penalty function defined as the sum of $P_c$ and $P_{tw}$, the penalty terms for the violation of the capacity and time window constraints, respectively, and $\alpha$ is a penalty coefficient whose value is adapted iteratively during the search process. $P_c$ is defined as the sum of the total excess of capacities in all routes. $P_{tw}$ is defined as $P_{tw} = \sum_{r=1}^m TW_r$, where $TW_r$ is the total of time window violation in the route $r$.

In the beginning, $\alpha$ is set to $1.0$. Then, if the second stage fails, $\alpha$  is updated. If $P_c < P_{tw}$, $\alpha$ is divided by $0.99$ to emphasize the penalty term of the time window constraint. Otherwise, $\alpha$ is multiplied by $0.99$.

After accepting an infeasible insertion that minimizes $F_p$, a series of local search moves using the local operator 2-opt$^*$, intra- and inter-route relocation, and intra- and inter-route exchange are performed to restore the feasibility of the partial solution. If the feasibility of the partial solution is restored, then we update the partial solution before selecting a new customer from $EP$. Otherwise, we move forward to the next stage.

Finally, since $v_{in}$ could not be inserted easily, its penalty value is updated: $p[v_{in}] = p[v_{in}] + 1$. Now, all partial solutions obtained by inserting $v_{in}$ at some position $p$ in some route $r$ of $\sigma$, and ejecting at most $k_{max}$ clients from $r$ are evaluated. Let $\mathcal{N}_{EJ}^{fe}(v_{in},\sigma)$ be the set of feasible partial solutions that are obtained in this way, ejecting customers $v_{out}^{(1)},...,v_{out}^{(k)}$ ($k \leq k_{max}$). Note that also $v_{in}$ itself can be ejected. The next solution is selected from $\mathcal{N}_{EJ}^{fe}(v_{in},\sigma)$, such that the sum of penalty counters of the ejected customers, $P_{sum} = p[v_{out}^{(1)}] + ... + p[v_{out}^{(k)}]$, is minimized. This criterion is motivated by the expectation that ejecting the customers whose sum of penalty counters is the smallest possible will benefit the subsequent insertions, even if the number of the customers in the ejection pool is momentarily increased. The ejected customers are added to $EP$, and the resulting feasible partial solution $\sigma$ is modified by a procedure of diversification, by executing moves selected at random from $\mathcal{N}(\sigma)$ for a given number of times. $\mathcal{N}(\sigma)$ is a composite neighborhood defined as the set of the partial solutions that are obtained by applying the move operators 2-opt$^*$, intra- and inter-route relocation, and intra- and inter-route exchange. At the end of this third stage, we update the partial solution before proceeding to select a new customer from $EP$.

\subsubsection{Multiperiod Route Elimination Heuristic for ConVRPTW}

\begin{figure}[th!]
  \centerline{\includegraphics[scale=0.7]{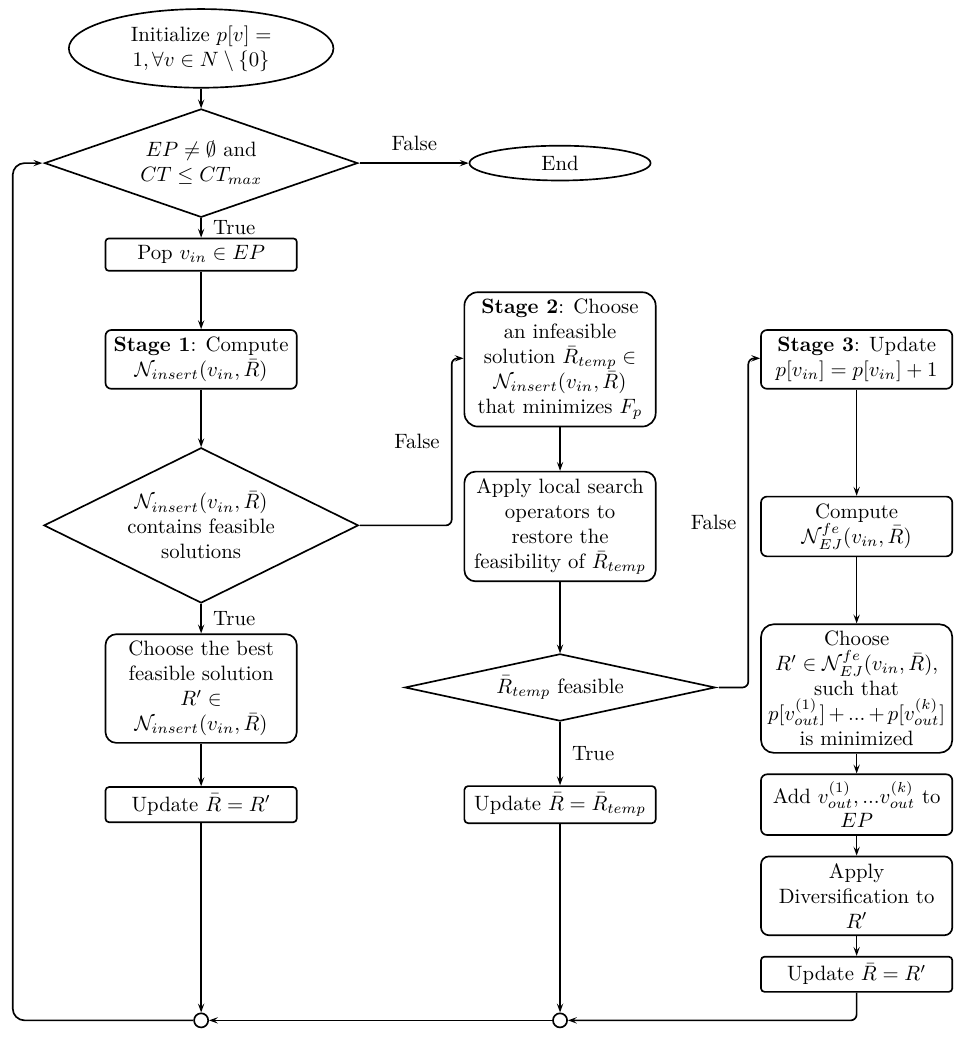}}
\caption{Multiperiod Route Elimination Heuristic for ConVRPTW\label{fig:3}}
\end{figure}

The multi-period route elimination heuristic for ConVRPTW requires as input a feasible initial solution $R$, which is obtained with the multi-period extension of the initial solution construction heuristic for ConVRPTW. Like in the one-period heuristic, stages 1 through 3 are repeatedly applied to the incumbent solution $R$ to reduce the number of vehicles one by one, until the total computation time reaches a given limit $CT_{max}$. Finally, we obtain a new solution $R$ with fewer drivers than the initial solution. See Figure \ref{fig:3} for a general overview.

Unlike the original one-period heuristic, the vehicle to eliminate is not chosen at random, but we select the vehicle with the least total number of {\em active} customer-days over the planning horizon (the sum over all customers of all days when they are active). The original idea to choose a vehicle at random proved not to be as effective, given the complexity of the instances that we need to deal with. We temporarily remove the vehicle, and we initialize the stack $EP$ with its customers. Then, a repeated attempt is made to insert the customers from the $EP$ into the $|K| - 1$ remaining subsets of customers, avoiding capacity and time window constraint violations, for each day that a selected customer is {\em active}.

In the first stage, we define the set of partial solutions that are obtained by inserting $v_{in}$ into all vehicles at all possible positions as $\mathcal{N}_{insert}(v_{in},R)$. Now, the insertion of a client into a vehicle considers an insertion position for each day (route) of the planning horizon. Insertion into a vehicle is feasible only if the corresponding insertion positions lead to feasible routes over the whole planning horizon. The quality of a feasible insertion into a vehicle is computed as the sum of the respective qualities over all days of the planning horizon (using the same formulas as in the multi-period initial solution construction). If there exist feasible insertions into a vehicle, then the next solution is not chosen at random like in the original one-period algorithm, but the best quality insertion into a vehicle is selected.

In the second stage, we proceed similar to the one-period heuristic, i.e. accepting temporarily an infeasible insertion into a vehicle that minimizes $F_p$, but extended for the multi-period case: $F_p(R) = \sum _{d\in D} P_{c,d}(\sigma) + \alpha P_{tw,d}(\sigma)$. Then, a series of local search moves, 2-opt, and inter-route relocation are  applied alternately to restore the feasibility of the partial solution. As in other cases of using these operators in our algorithm, we apply the best fit strategy. Not all local search operators used in \cite{nagata2009} are adopted here due to the complexity of working with multiple periods. If the feasibility of the partial solution is restored, then we update the partial solution before proceeding to select a new customer from $EP$. Otherwise, we move forward to the next stage.

Finally, in the third stage, since $v_{in}$ could not be inserted easily, its penalty value is updated: $p[v_{in}] = p[v_{in}] + 1$. Then, in contrast to the one-period heuristic, we define $\mathcal{N}_{EJ}^{fe}(v_{in},R)$ as the set of feasible partial solutions that are obtained by inserting $v_{in}$ into a vehicle that is chosen at random at positions that minimize the value of $F_p$, and ejecting at most $k_{max}$ customers, $v_{out}^{(1)},...,v_{out}^{(k)}$ ($k \leq k_{max}$) from the resulting infeasible vehicle in all possible ways. Here also $v_{in}$ itself can be ejected. The next solution is selected from $\mathcal{N}_{EJ}^{fe}(v_{in},R)$ such that the sum of the penalty counters of the ejected customers, $P_{sum} = p[v_{out}^{(1)}] + ... + p[v_{out}^{(k)}]$, is minimized. The ejected customers are added into $EP$, and the resulting feasible partial solution $R$ is modified by a procedure of diversification. This diversification process works on the objective to minimize the total travel distance for an incomplete solution, that is different from the main algorithm which aims to minimize the number of routes for the full solution, and consists of applying repeteadly inter-route relocation and intra-route 2-opt local operators until no improvement is found. Then, we update the current partial solution before proceeding to select a new customer from $EP$.

\section{Experimental Results}
The proposed algorithm was implemented in Python (see \cite{R07}), using NumPy (see \cite{O15}) and Numba (see \cite{LPS2015}). We tested the proposed algorithm on sets of small and large instances. Small instances were constructed from Solomon's VRPTW instances and were solved optimally with Gurobi 9.0.1 on an Intel Core i7 with 2.67GHz$\times$8 and 4GiB of memory. Large instances were constructed from the data of the Chilean company for a year planning horizon. Time and distance matrices were computed using Open Source Routing Machine software (see \cite{LV2011}) based on Open Street Maps data.

\subsection{Instance Characteristics}

\subsubsection{Small ConVRPTW Instances}
We generate 56 small instances from VRPTW benchmark instances of \cite{solomon1987}. Solomon's instances are divided into six classes. Classes C1 and C2 have clustered geographical distribution of customers, in classes R1 and R2 the geographical distribution is randomly uniform, and classes RC1 and RC2 have a mix of both geographically uniform and clustered customers. Classes R1, C1 and RC1 have narrow time windows and small vehicle capacities of 200, so the number of customers served by a single vehicle is small. Classes R2, C2 and RC2 have wide time windows over longer scheduling horizons and large vehicle capacities of 700 for C2 and 1000 for R2 and RC2, so the solutions have comparatively fewer vehicles. All of the classes consider 100 customers and a single depot.

From Solomon's instances, we construct small ConVRPTW instances. We consider only the first ten customers, and we keep without modifications the customers' demand, coordinates, time windows, and service times. We consider a planning horizon of five days. The service frequency, i.e., probability that a customer requires service on a given day, was set to 70\%, as in \cite{groer2009}. We reduced the capacity of vehicles by half.

\subsubsection{Large ConVRPTW Instances}
A set of thirteen instances was constructed from the company's historical demand data from January 2018 until March 2019, based on moving trimesters (first starting in January 2018 and last in January 2019). Each one of them is characterized by a different number of customers and days. The vehicle fleet is homogeneous, with vehicles of 5 tons each.

In Table \ref{tab:1} we present a brief description of these instances. They are characterized by a long planning horizon of 71 days, 2115 customers, and 24163 orders, on average. Moving from one moving trimester to the next one, on average, there are 3.4\% of new customers. We highlight that, on average, only 16.1\% of the customers are active each day, a low proportion of the total number of customers. This situation makes it particularly difficult to design consistent routing plans.

\begin{table}[th!]
\caption{Trimestral instances}\label{tab:1}
\begin{tabular*}{\hsize}{@{}@{\extracolsep{\fill}}llcccccccc@{}}\toprule
&&&&\multicolumn{2}{c}{\textbf{Customers}}&\multicolumn{4}{c}{\textbf{Active customers per day [\%]}}\\
\cmidrule(r){5-6}\cmidrule(r){7-10}
Id  &Instance &Days &Orders  &Current 	&New[\%] &Mean &Std. &Min. &Max. \\
\toprule

1	&January2018-March2018			&65		&24525		&2366	&-		&15.9 &3.4	&6.7  &21.6 \\
2	&February2018-April2018			&64		&23064		&2333	&4.6	&15.4 &3.3	&7.6  &21.6 \\
3	&March2018—May2018				&62		&22039		&2254	&3.3	&15.8 &3.4	&6.3  &21.0 \\
4	&April2018-June2018				&70		&24688		&2187	&3.5	&16.1 &3.3	&6.5  &21.2 \\
5	&May2018-July2018				&71		&24659		&2143	&2.7	&16.2 &3.6	&6.6  &21.7 \\
6	&June2018-August2018			&79		&27303		&2140	&3.1	&16.1 &3.3	&6.8  &22.1 \\
7	&July2018-September2018			&75		&25412		&2081	&2.0	&16.3 &3.7	&6.5  &22.7 \\
8	&August2018-October2018			&73		&24211		&2060	&2.3	&16.1 &3.7	&5.7  &23.0 \\
9	&Septermber2018-Novermber2018	&70		&22012		&2020	&3.9	&15.6 &3.9	&5.8  &22.6 \\
10	&October2018-December2018		&73		&22908		&2013	&4.4	&15.6 &3.7	&5.9  &23.2 \\
11	&Novermber2018-January2019		&75		&24060		&2006	&5.0	&16.0 &3.4	&6.7  &23.3 \\
12	&December2018-February2019		&75		&24834		&1963	&2.2	&16.9 &3.0	&9.1  &23.8 \\
13	&January2019-March2019			&74		&24405		&1941	&3.7	&17.0 &2.5	&11.5 &21.6 \\
	&Average						&71.2	&24163.1	&2115.9	&3.4	&16.1 &3.4	&7.1  &22.3 \\
\bottomrule
\end{tabular*}
\end{table}

In Table \ref{tab:2}, we present in a similar way a set of fifteen instances built on individual months.

\begin{table}[th!]
\caption{Monthly instances}\label{tab:2}
\begin{tabular*}{\hsize}{@{}@{\extracolsep{\fill}}llccccccccc@{}}\toprule
&&&&\multicolumn{2}{c}{\textbf{Customers}}&\multicolumn{4}{c}{\textbf{Active customers per day [\%]}}\\
\cmidrule(r){5-6}\cmidrule(r){7-10}
Id&Instance	&Days &Orders  &Current &New[\%] &Mean &Std. &Min. &Max. \\
\toprule

1	&January2018	&26		&10142	&2079	&-		&18.8	&4.1	&7.6	&24.6 \\
2	&February2018	&21		&7857	&1958	&10.0	&19.1	&4.0	&12.6	&25.8 \\
3	&March2018		&18		&6526	&1840	&12.1	&19.7	&4.2	&9.6	&25.7 \\
4	&April2018		&25		&8681	&1892	&16.1	&18.4	&3.7	&11.5	&23.6 \\
5	&May2018		&19		&6832	&1776	&9.7	&20.2	&4.5	&8.0	&26.1 \\
6	&June2018		&26		&9175	&1899	&15.3	&18.6	&3.5	&12.8	&23.1 \\
7	&July2018		&26		&8652	&1772	&8.2	&18.8	&4.6	&8.2	&26.1 \\
8	&August2018		&27		&9476	&1859	&12.6	&18.9	&3.4	&12.5	&25.4 \\
9	&September2018	&22		&7284	&1719	&7.6	&19.3	&5.0	&7.9	&26.6 \\
10	&October2018	&24		&7451	&1698	&11.3	&18.3	&4.2	&6.9	&24.3 \\
11	&November2018	&24		&7277	&1641	&12.0	&18.5	&4.5	&8.2	&26.1 \\
12	&December2018	&25		&8180	&1678	&13.1	&19.5	&4.4	&10.6	&27.8 \\
13	&January2019	&26		&8603	&1714	&12.8	&19.3	&2.8	&13.1	&23.5 \\
14	&February2019	&24		&8051	&1622	&8.0	&20.7	&3.0	&15.4	&25.8 \\
15	&March2019		&24		&7751	&1625	&12.0	&19.9	&3.1	&14.5	&25.5 \\
	&Average		&23.8	&8129.2	&1784.8 &11.5	&19.2	&3.9	&10.6	&25.3 \\
\bottomrule
\end{tabular*}
\end{table}

\subsection{Experimental Settings}

The parameters to be adjusted in our algorithm correspond to the constructive and the route elimination heuristics.
To get the values for the constructive heuristic, first we solved ConVRP instances from \cite{groer2009} and chose the values that gave the best initial solution considering the total number of vehicles and total travel time overall the instances. We got the following values: $ic = 1$, $\mu = 1$, $\lambda = 2$, $\alpha_1 = 0.5$ and $\alpha_2 = 0.5$. Later on, we tried if modifying these values led to better results on the instances coming from our case study, but no significant improvement could be obtained.

For the route elimination heuristics we used the values of parameters proposed in \cite{nagata2009} for the maximum computational time $CT_{max} = \{1,10,60\}$, in minutes. And we set $k_{max} = 3$.

The values of these parameters were maintained unchanged for all the instances. Consistent good performance of these parameter values makes our algorithm easy to implement since no parameter tuning is necessary.

\subsection{Results for Small ConVRPTW Instances}
We evaluate the performance of the ConVRPTW algorithm by comparing the solutions obtained by ConVRPTW with the optimal solution found by Gurobi, with a time limit of 1 hour. All the instances were solved optimally in 1 hour, except one instance from the class R1, namely the R102 instance, with GAP of 33.3\%.

The results of the comparison are reported in Table \ref{tab:3g}. We present the average result for every class. In the first three columns, we report the optimal solution with respect to the average number of vehicles (ANV), average total computational time (ACPU) in seconds, and average gap (AGAP) in percentage. In the following two columns, we report the solution found by ConVRPTW algorithm with respect to ANV and ACPU. Finally, in the last column, we report the percentage difference in the number of vehicles $\Delta_{NV}$ between the best known and the ConVRPTW solution.

We can see that the number of vehicles obtained by the ConVRPTW algorithm for each class is the same as the one obtained with the MILP model using Gurobi, but with a much lower computing time. This shows the effectiveness of the ConVRPTW algorithm in minimizing the number of vehicles. Finally, we comment that for Instance R102 from class R1, we only know a solution with a GAP of 33.3\%, and in this case, the ConVRPTW algorithm finds a solution with the same number of vehicles.

\begin{table}[th!]
\caption{Results for small ConVRPTW instances}\label{tab:3g}
\begin{tabular*}{\hsize}{@{}@{\extracolsep{\fill}}lccclccc@{}}\toprule
&\multicolumn{3}{c}{\textbf{Gurobi}}&\multicolumn{2}{c}{\textbf{ConVRPTW}}\\

\cmidrule(r){2-4}\cmidrule(r){5-6}
Instance	&ANV	&ACPU &AGAP	&ANV	&ACPU &$\Delta_{NV}$\\
  			&  	&[s] &[\%]	&    &[s] &[\%]
 \\\midrule

C1		&1.9	&4.4	&0.0	&1.9	&0.9	&0.0 \\ 	
C2		&1.0	&2.4	&0.0	&1.0	&0.0	&0.0 \\
R1		&2.3	&458.3	&2.8	&2.3	&1.0	&0.0 \\
R2		&1.0	&5.1	&0.0	&1.0	&0.0	&0.0 \\
RC1		&2.9	&32.0	&0.0	&2.9	&1.0	&0.0 \\
RC2		&1.0	&4.5	&0.0	&1.0	&0.0	&0.0 \\
Avg.	&1.7	&84.4	&0.5	&1.7	&0.5	&0.0 \\
\bottomrule
\end{tabular*}
\end{table}

\subsection{Results for Large ConVRPTW Instances}
The performance of the proposed algorithm for ConVRPTW is compared with the current routing plan used by the company. The company used a one-to-one client-to-driver assignment to execute the delivery of the orders. Because of the requirement of client-to-driver consistency, knowledge of the orders only one day in advance, and the variability of the active customers every day, it is common that the current routing plan in use by the company has many customers for which the time windows are not respected. We evaluate the performance of the current routing plan of the company in terms of the number of vehicles used for executing the deliveries (NV), the resulting total travel time of the complete routing plan (TT) in minutes, the percentage of visits for which the time windows are not respected (PTW), and the lateness of deliveries with respect to TWs as the proportion of TT (LTW).

In tables \ref{tab:3} and \ref{tab:4} we describe the performance of the current operation of the company on the trimestral and monthly group of instances, respectively, in contrast to the results generated by our algorithm for ConVRPTW. In the first three columns of these tables, we report NV, TT, and PTW of the current routing plan of the company. In the next columns, we report the results obtained by applying the proposed algorithm to ConVRPTW instances, which are evaluated by NV, the percentage of improvement of the total travel time $\Delta_{TT}$, and the total computational time (CPU) in seconds. The last column gives the percentage of improvement in the number of vehicles $\Delta_{NV}$ between the current solution of the company and the best results obtained with the ConVRPTW algorithm. We present the results obtained using maximum computation times of 1, 10, and 60 minutes used for our multi-period route elimination heuristic.
The execution time of the complete program is equal to the chosen configuration time plus the time spent for the multi-period initial solution construction heuristic and the final re-optimization heuristic.

For the group of trimestral instances, in Table \ref{tab:3}, analyzing the performance of the current routing plan of the company, we can see that the number of vehicles required for executing the deliveries ranges from 21 to 25 vehicles, there exists an average of 18.4\% of customers for which the time windows are not respected, and the average of the lateness of deliveries with respect to TWs as the proportion of the TT is 4.8\%. The total travel time required for the complete routing plan is, on average, 12630.9 hours. For this group of instances, our algorithm reaches solutions with a low number of vehicles already with the setting of 1 minute. Moreover, the total travel times are improved by 8\%, on average, with respect to the company's routing plan, and the customers' time windows are fully respected. Using the settings of 10 minutes leads to a further reduction of one vehicle for instance 12. In the case of 60 minutes, a solution with one vehicle less is found not only for instance 12 but also for instances 1 and 11. The total travel times do not change, on average, from one time setting to another. The percentage of improvement in the number of vehicles is, on average, 31.7\%. In brief, the company's routing plan is improved, reducing the number of vehicles required for the deliveries, on average, in seven vehicles. The total travel times are improved by 8\%, on average, and the customers are serviced within their time windows.

\begin{table}[th!]
\caption{Results for trimestral instances }\label{tab:3}
\begin{tabular*}{\hsize}{@{}@{\extracolsep{\fill}}lcccccccccccccc@{}}\toprule
&\multicolumn{4}{c}{\textbf{Current Solution}}&\multicolumn{9}{c}{\textbf{ConVRPTW}} \\
 \cmidrule(r){2-5} \cmidrule(r){6-14}
&&&&& &1min. && &10min. && &60min.& &\\

\cmidrule(r){2-5}\cmidrule(r){6-8}\cmidrule(r){9-11} \cmidrule(r){12-14}
Id.&NV &TT 		&PTW &LTW	&NV   &$\Delta_{TT}$  &CPU &NV &$\Delta_{TT}$  &CPU &NV &$\Delta_{TT}$    &CPU  &$\Delta_{NV}$\\
  &   &[h]      &[\%] &[\%]	&    &[\%]&[s] &   &[\%]&[s] &  &[\%]&[s] & [\%]\\
\toprule
1	&23		&12204.5	&26.8	&8.8	&17		&7.6	&124	&17		&7.6	&993	&16		&8.3	&3769	&30.4 \\	
2	&23		&11596.0	&17.4	&5.4	&16		&6.0	&115	&16		&5.3	&635	&16		&6.3	&3784	&30.4 \\	
3	&25		&11163.4	&18.5	&6.0	&16		&8.4	&97		&16		&8.4	&635	&16		&8.4	&3696	&36.0 \\	
4	&24		&12598.4	&21.2	&5.9	&16		&7.0	&129	&16		&7.0	&689	&16		&7.0	&3674	&33.3 \\	
5	&22		&12874.2	&24.9	&6.7	&16		&7.2	&103	&16		&7.2	&652	&16		&7.2	&3650	&27.3 \\	
6	&22		&14264.4	&20.2	&4.9	&16		&9.1	&347	&16		&9.1	&650	&16		&9.1	&3718	&27.3 \\	
7	&23		&13516.4	&17.3	&4.2	&16		&9.3	&115	&16		&9.3	&697	&16		&9.3	&3690	&30.4 \\	
8	&23		&12804.1	&13.6	&3.3	&16		&9.3	&552	&16		&9.3	&684	&16		&9.3	&3651	&30.4 \\	
9	&23		&11982.1	&14.1	&3.6	&16		&6.6	&356	&16		&6.6	&777	&16		&6.6	&3704	&30.4 \\	
10	&24		&12346.6	&13.7	&3.1	&15		&8.6	&99		&15		&9.1	&655	&15		&8.6	&3651	&37.5 \\	
11	&23		&12894.7	&14.7	&3.1	&16		&7.4	&343	&16		&7.4	&624	&15		&7.5	&3689	&34.8 \\	
12	&23		&13125.2	&18.8	&4.1	&16		&9.3	&295	&15		&9.8	&641	&15		&8.1	&3983	&34.8 \\	
13	&21		&12831.5	&17.8	&4.0	&15		&9.3	&394	&15		&9.3	&838	&15		&9.3	&3842	&28.6 \\	
Avg	&23.0	&12630.9	&18.4	&4.8	&15.9 	&8.1	&236	&15.8	&8.1	&705	&15.7	&8.1	&3731	&31.7 \\

\bottomrule
\end{tabular*}
\end{table}

For the group of monthly instances, in Table \ref{tab:4}, in the current routing plan, the number of vehicles required for executing the deliveries ranges from 21 to 23 vehicles, there exists an average of 7.6\% of customers for which the time windows are not respected, and an average of the lateness of deliveries with respect to TWs as the proportion of TT of 5.1\%. The total travel time required for the complete routing plan is, on average, 4233.2 hours. For this group of instances, our algorithm also reaches good solutions with the setting of 1 minute. Using the setting of 10 minutes leads to a further reduction of one vehicle on instances 6,7,8,11,12, and 60 minutes gives one vehicle less on almost all of the instances. The total travel times at most range, on average, 0.04\% from one time setting to another. The percentage of improvement in the number of vehicles is, on average, 35.5\%. In brief, the routing plan of the company is improved reducing the number of vehicles required for the deliveries, on average, in six vehicles, the total travel times are improved by 10\%, on average, and the customers are serviced within their time windows.

\begin{table}[th!]
\caption{Results for monthly instances}\label{tab:4}
\begin{tabular*}{\hsize}{@{}@{\extracolsep{\fill}}lcccccccccccccc@{}}\toprule
&\multicolumn{4}{c}{\textbf{Current Solution}}&\multicolumn{9}{c}{\textbf{ConVRPTW}} \\
 \cmidrule(r){2-5} \cmidrule(r){6-14}
&&&&& &1min. && &10min. && &60min.& &\\

\cmidrule(r){2-5}\cmidrule(r){6-8}\cmidrule(r){9-11} \cmidrule(r){12-14}
Id.&NV &TT 		&PTW &LTW	&NV   &$\Delta_{TT}$  &CPU &NV &$\Delta_{TT}$  &CPU &NV &$\Delta_{TT}$    &CPU  &$\Delta_{NV}$\\
  &   &[h]      &[\%] &[\%]	&    &[\%]&[s] &   &[\%]&[s] &  &[\%]&[s] & [\%]\\
\toprule
1	&22		&5024.5	&16.4	&12.5	&16		&10.4	&109	&16		&10.4	&717	&16		&10.4	&3616	&27.3 \\
2	&22		&3901.1	&8.3	&6.0	&16		&9.2	&122	&16		&9.2	&626	&15		&10.0	&3627	&31.8 \\
3	&23		&3278.8	&7.1	&6.5	&15		&9.4	&69		&15		&9.4	&614	&14		&7.6	&3611	&39.1 \\
4	&22		&4416.1	&6.0	&4.2	&14		&10.5	&149	&14		&11.2	&672	&13		&10.1	&3667	&40.9 \\
5	&22		&3468.5	&9.8	&7.7	&15		&10.9	&82		&15		&10.9	&619	&14		&8.0	&3610	&36.4 \\
6	&22		&4713.9	&9.3	&6.3	&15		&10.5	&100	&14		&10.5	&627	&14		&11.1	&3625	&36.4 \\
7	&22		&4691.8	&10.3	&6.4	&16		&9.7	&109	&15		&10.6	&632	&15		&10.7	&3614	&31.8 \\
8	&22		&4858.7	&3.9	&2.1	&15		&10.6	&126	&14		&12.4	&646	&14		&12.7	&3614	&36.4 \\
9	&23		&3965.9	&6.1	&4.4	&15		&11.3	&75		&15		&11.3	&622	&15		&11.3	&3625	&34.8 \\
10	&22		&3979.6	&6.1	&3.7	&14		&10.7	&115	&14		&10.7	&657	&14		&10.7	&3643	&36.4 \\
11	&21		&4036.7	&4.7	&2.6	&15		&10.8	&115	&14		&12.0	&608	&13		&9.0	&3606	&38.1 \\
12	&23		&4330.3	&5.7	&2.9	&15		&9.5	&130	&14		&10.0	&615	&14		&10.1	&3607	&39.1 \\
13	&21		&4527.7	&7.2	&3.7	&14		&13.6	&89		&14		&13.6	&626	&13		&8.4	&3610	&38.1 \\
14	&21		&4267.2	&9.3	&5.7	&14		&13.3	&73		&14		&13.3	&630	&14		&13.3	&3625	&33.3 \\
15	&21		&4036.6	&4.4	&2.5	&14		&12.8	&92		&14		&12.8	&621	&14		&12.8	&3607	&33.3 \\
Avg	&21.9	&4233.2	&7.6	&5.1	&14.9	&10.9	&104	&14.5	&11.2	&635	&14.1	&10.4	&3620	&35.5 \\
\bottomrule
\end{tabular*}
\end{table}

Both for trimestral and monthly instances, the algorithm yields good results already with 1 minute of computation. Indeed, increasing the computational times to 10 or 60 minutes only leads to minor improvements in terms of the number of vehicles (NV).

Notice that, since the total travel time (TT) is not part of the objective function that we are optimizing, in some cases more computational time leads to higher values of this attribute. This is partly due to the restrictions imposed by time windows, where reducing the number of vehicles may force longer routes - like in the monthly instance number 3, 4, 5, 11, 13, when augmenting the computation time from 10 to 60 minutes. But it may also be due to certain random characteristics of our algorithm - like in the trimestral instances number 10 and 12, when moving from 10 to 60 minutes.

\subsection{Driver assignment (operational) management}
In this subsection, we provide some managerial insights by evaluating the results of the proposed heuristic in the context of the operational reality of the company under study. With a fixed client-to-driver assignment, the company tries to satisfy the customers' demand for every unknown operational day. However, because of the variability of customers' demand and activity, the company is forced to implement ad-hoc changes to the assignment on individual days. For instance, when on a particular day, a driver has active customers with a demand that exceeds the vehicle's capacity, then some of these customers are reassigned to other drivers. In some cases, there may even be orders not fulfilled at all. Therefore, this situation generates delays in the deliveries and reduces the consistency of the service.

The general idea is to use the company's historical data to compute the initial client-to-driver assignment and use it for the daily operational routing during one month. After that, the assignment should be updated each month, based on the updated historical data. We experiment with two settings. The first option is to create the initial assignment based on the last three months of operational history and then, each month, do an update to the assignment based on the recent moving trimester. The second option is similar, only that we base the initial solution and the monthly updates on the recent month.

Our analysis begins with the ConVRPTW solutions of the first trimestral and monthly instances, respectively. Then, we choose a period of one-month to update the ConVRPTW solution recursively, and we analyze its daily performance during the following month. Our objective is to evaluate how client-to-driver consistency is affected in order to satisfy the daily demand for a new unknown period (month).

First, we compute the initial ConVRPTW solutions for the first trimestral and monthly past data instance, respectively. Then, we develop two kinds of computations: i) updating the previous ConVRPTW solution based on the following month (the following moving trimester or just the following month instance, respectively) and (ii) operational daily routing for each day of the following month.

\subsubsection{Results for Updating ConVRPTW Solutions}
The experiments that update a ConVRPTW solution start with a ConVRPTW solution for one period to update it based on the customers' demand for the following period. A ConVRPTW solution for the following moving trimester or month is computed, based on the initial solution obtained for the previous moving trimester or month. This problem corresponds to the situation at a company, where each month the customers' demand and activity during the most recent period is analyzed, and the initial client-to-driver assignment can be updated if necessary.

For example, for trimestral instances, based on the ConVRPTW solution of Instance 1, which corresponds to January, February, and March, we compute a solution of Instance 2, comprising February, March, and April. Note that now, the solution of Instance 2 is not obtained by solving the instance from scratch, but by updating the solution of Instance 1, which is a heuristic procedure to minimize the change in the client-to-driver assignment.

Therefore, to create a ConVRPTW solution for the following instance, we use the solution of the previous instance as the initial solution for our multi-period route elimination heuristic. We put all the new customers on the stack $EP$, delete the old customers with no demand in the current period, and run the rest of the algorithm as we did for independent ConVRPTW instances. With the solution of the new instance, we evaluate how the consistency of the old customers is affected: how many clients had their driver changed in the update.

In Table \ref{tab:5}, we report the results of the experiments on the groups of trimestral and monthly instances. In the first four columns, we report the results of working recursively on the trimestral instances (starting from Trimestral Instance 1), and in the following columns the results for the monthly instances (starting from Monthly Instance 1). So, in both cases, the solution from a previous instance is used as the initial solution for the following instance.

For the group of trimestral instances, we can note that when using the client-to-driver assignment obtained on Instance 1 as the initial solution for Instance 2, it is necessary to use an additional vehicle. The set of vehicles is maintained in posterior updates. Also, we can see that the percentage of inconsistent customers (IC), i.e., the percentage of customers assigned to a different driver during the update is 7.8\%, on average.

For the group of monthly instances, we can note that it is unnecessary to use a new vehicle to update the ConVRPTW solution from Instance 1 to Instance 2. The percentage of inconsistent customers is, on average, 9\%.

\begin{table}[h]
\caption{Results for updating ConVRPTW solutions experiments for moving trimesters and months}\label{tab:5}
\begin{tabular*}{\hsize}{@{}@{\extracolsep{\fill}}lccc|lccc@{}}\toprule
\multicolumn{4}{c}{\textbf{Updated moving trimester}}&\multicolumn{4}{c}{\textbf{Updated moving month}}\\

	\cmidrule(r){1-4}		\cmidrule(r){5-8}
Id	&NV	&TT			&IC		&Id		&NV		&TT			&IC		\\
	&	&[h]     &[\%]	&		&		&[h]     &[\%]
 \\\midrule
1	&16	& 11188.5 &-		&1	&16 &4504.0 &-		\\ 	
2	&17	& 11117.2 &8.6		&2	&16 &3606.6 &15.8	\\	
3	&17	& 10923.9 &14.5		&3	&16 &3043.6 &10.8	\\	
4	&17	& 12258.1 &8.3		&4	&16 &4137.1 &11.1	\\	
5	&17	& 12941.9 &10.3		&5	&16 &3178.5 &9.2	\\	
6	&17	& 13709.0 &8.4		&6	&16 &4304.2 &11.8	\\	
7	&17	& 12910.6 &6.0		&7	&16 &4289.1 &6.3 	\\	
8	&17	& 15861.8 &10.1		&8	&16 &4477.3 &5.4	\\	
9	&17	& 15371.4 &10.8		&9	&16 &3619.3 &12.0	\\	
10	&17	& 11540.6 &5.8		&10	&16 &3648.5 &7.4	\\	
11	&17	& 11919.3 &1.4		&11	&16 &3724.6 &6.6	\\	
12	&17	& 12129.3 &9.3		&12	&16 &4025.8 &12.6	\\	
13	&17	& 11082.1 &0.3		&13	&16 &4072.2 &5.0	\\	
	&	&	      &			&14	&16 &3810.8 &6.1	\\		
	&	&	      &			&15	&16	&3634.7 &6.4	\\	

Average	&16.9	&12534.9 &7.8	&Average &16	&3871.7	&9.0	\\
\bottomrule
\end{tabular*}
\end{table}

\subsubsection{Results for Operational Routing the Following Month}
Experiments of the operational routing during the following month use the client-to-driver assignment obtained by updating ConVRPTW on the most recent historical period. This assignment is used to respond to the customers' orders on individual days during the following month (the customers' demand and activity for which is not known a priori).

For example, based on the ConVRPTW solution of Trimestral Instance 1, which corresponds to January, February, and March, we solve the operational routing problem for every day of the following month of April. Note that the behavior of old customers in April is different than it was during the recent period (January-March), including that some of them will make no orders during the whole month. On the other hand, there will be some new customers, that presented no activity in the recent period.

The computational procedure we apply here is similar to that of the previous subsection, only that now we compute the routing for each day independently. We use the last available ConVRPTW solution as the initial solution for the current day, putting all new customers into the stack $EP$, and deleting the customers with no demand. Finally, we apply the route elimination heuristic to find the daily operational routing plan. Then, we evaluate how the consistency of the old customers is affected on each day of the following month, with daily routing restricted to fulfill all the orders within their respective time windows.

In Table \ref{tab:6}, we report the results of the experiments on the groups of trimestral and monthly instances. For each instance, we use the updating ConVRPTW solutions described in the previous subsection.

For the group of trimestral instances, we report the results of the following operational month experiments. We can see that the levels of inconsistency, the percentages of visits that do not respect the base client-to-driver assignment, are quite low reaching an average of 0.8\%.

For the group of monthly instances, we can see that the levels of inconsistency are still quite low: 1.1\%, on average. Therefore, a situation that requires not to respect the base client-to-driver assignment happens rarely.

\begin{table}[h]
\caption{Results for operational following month experiments from trimestral and monthly instances}\label{tab:6}
\begin{tabular*}{\hsize}{@{}@{\extracolsep{\fill}}lcccc|lcccc@{}}\toprule
\multicolumn{5}{c}{\textbf{Operational following month from trimesters}}&\multicolumn{5}{c}{\textbf{Operational following month from months}}\\

\cmidrule(r){1-5}						\cmidrule(r){6-10}
Id	&Following month &NV &TT &IC 	&Id	&Following month	&NV	&TT		&IC\\
	&			&   &[h]		&[\%]	&	&			&   &[h]		&[\%]
 \\\midrule
1	&April		&16	&4137.2 &1.9     &1		&February2018	&16	&3734.8 &5.2 \\
2	&May		&17	&3243.4 &1.2     &2		&March2018		&16	&3076.3 &1.4 \\
3	&June		&17	&4391.0 &0.8     &3		&April2018		&16	&4135.0 &1.3 \\
4	&July		&17	&4370.8 &0.7     &4		&May2018		&16	&3182.8 &1.1 \\
5	&August		&17	&4630.5 &0.5     &5		&June2018		&16	&4325.3 &0.9 \\
6	&September	&17	&3670.6 &0.7     &6		&July2018		&16	&4226.4 &0.6 \\
7	&October	&17	&3814.2 &0.2     &7		&August2018		&16	&4425.9 &0.6 \\
8	&November	&17	&3811.8 &1.1     &8		&September2018	&16	&3579.8 &1.2 \\
9	&December	&17	&4072.0 &1.8     &9		&October2018	&16	&3695.3 &0.4 \\
10	&January	&17	&4145.1 &0.1     &10	&November2018	&16	&3693.0 &0.3 \\
11	&February	&17	&3907.9 &0.3     &11	&December2018	&16	&4023.2 &1.6 \\
12	&March		&17	&3654.0 &0.0     &12	&January2019	&16	&4053.4 &0.5 \\
	&			&	&	    &        &13	&February2019	&16	&3811.8 &0.5 \\
	&			&	&       & 		 &14	&March2019		&16	&3629.1 &0.3 \\

Average	& &16.9 &3987.4	&0.8         &Average &			&16	&3828.0	&1.1\\
\bottomrule
\end{tabular*}
\end{table}

\section{Conclusions}

Classical studies on vehicle routing tend to focus on finding routing plans of minimum cost, presenting methods for solving the standard one-period Vehicle Routing Problem with Time Windows (VRPTW). However, in competitive markets, it is important to find a balance between operational cost and customer satisfaction to assure long term profitabilities. In this work, we consider the Consistent Vehicle Routing Problem with Time Windows (ConVRPTW), an extension of the standard VRPTW to multiple periods, that considers the traditional constraints on the vehicle capacity and time windows of the customers combined with the requirement of multi-period driver consistency. This problem is motivated by a real-world application at a food company's distribution center, in which the main objective is to design driver consistent routing plans with a minimum number of vehicles and total travel time as low as possible.

We develop a new heuristic approach to solve ConVRPTW based on the most successful algorithms in the literature for the one-period VRPTW. The performance of the proposed algorithm is evaluated on small instances generated from benchmark instances of the literature and instances constructed from the operational data of the company under study. The results show the effectiveness of the ConVRPTW algorithm in minimizing the number of vehicles. The proposed algorithm is capable of generating better consistent routing plans than the ones currently used by the company within low computational times. Additionally, we evaluate the solutions of ConVRPTW on historical data as the basis for the client-to-driver assignment for operational management under the conditions of customer activities and order quantities known only one day in advance. The results show that the proposed algorithm can be used to generate updated client-to-driver assignments that let the daily operational routing be solved with a very high driver consistency.

Considering the high variability in the activity of the customers, the quantities they order, and the service times, the proposed algorithm represents a basis for the development of a decision support system that would improve the tactical routing plans for serving the customers of the company, obtaining a good balance between operational costs and service quality.

Future work must be concentrated on further developing the proposed algorithm to output client-to-driver assignments that correspond to geographical zones that do not overlap. Such a characteristic would be very valuable for the management of the company, for the ease of establishing service contracts with the transportation companies based on general territories and not on particular sets of clients. This setting is easier to deal with under the high variability in client activity. For companies, this would represent a useful tool to determine the best way to serve and to respond quickly to new customer demands, standardize the quality of service, and make an efficient and equitable allocation of customers among their different drivers. 
Finally, we plan to embed the proposed algorithm in a multistart framework, like the one presented in \cite{marzo2020},
to improve the efficacy of exploration of the search space, thanks to stronger intensification and diversification strategies.
It might be necessary in order to obtain efficient algorithms for more constrained versions of the problem - for example,
with restrictions related to some geometrical properties of the geographical zones.

\section*{Acknowledgments}
H.L. gratefully acknowledges financial support from ANID + PAI/Concurso Nacional Tesis de Doctorado en el Sector Productivo, 2017 + Folio T7817120007. K.S. gratefully acknowledges financial support from Programa Regional STICAMSUD 19-STIC-05.


\bibliographystyle{acm}
\bibliography{references1}

\begin{thebibliography}{10}

\bibitem{braekers2016}
{\sc Braekers, K., and Kovacs, A.~A.}
\newblock A multi-period dial-a-ride problem with driver consistency.
\newblock {\em Transportation Research Part B: Methodological 94\/} (2016),
  355--377.

\bibitem{braysy2005a}
{\sc Br{\"a}ysy, O., and Gendreau, M.}
\newblock Vehicle routing problem with time windows, part i: Route construction
  and local search algorithms.
\newblock {\em Transportation science 39}, 1 (2005), 104--118.

\bibitem{campelo2019}
{\sc Campelo, P., Neves-Moreira, F., Amorim, P., and Almada-Lobo, B.}
\newblock Consistent vehicle routing problem with service level agreements: A
  case study in the pharmaceutical distribution sector.
\newblock {\em European Journal of Operational Research 273}, 1 (2019),
  131--145.

\bibitem{dantzig1959}
{\sc Dantzig, G.~B., and Ramser, J.~H.}
\newblock The truck dispatching problem.
\newblock {\em Management science 6}, 1 (1959), 80--91.

\bibitem{feillet2014}
{\sc Feillet, D., Garaix, T., Lehu{\'e}d{\'e}, F., P{\'e}ton, O., and Quadri,
  D.}
\newblock A new consistent vehicle routing problem for the transportation of
  people with disabilities.
\newblock {\em Networks 63}, 3 (2014), 211--224.

\bibitem{gendreau2010}
{\sc Gendreau, M., and Tarantilis, C.~D.}
\newblock {\em Solving large-scale vehicle routing problems with time windows:
  The state-of-the-art}.
\newblock Cirrelt Montreal, 2010.

\bibitem{goeke2019}
{\sc Goeke, D., Roberti, R., and Schneider, M.}
\newblock Exact and heuristic solution of the consistent vehicle-routing
  problem.
\newblock {\em Transportation Science 53}, 4 (2019), 1023--1042.

\bibitem{groer2009}
{\sc Gro{\"e}r, C., Golden, B., and Wasil, E.}
\newblock The consistent vehicle routing problem.
\newblock {\em Manufacturing \& service operations management 11}, 4 (2009),
  630--643.

\bibitem{hernandez2017}
{\sc Hernandez, F., Gendreau, M., and Potvin, J.-Y.}
\newblock Heuristics for tactical time slot management: a periodic vehicle
  routing problem view.
\newblock {\em International transactions in operational research 24}, 6
  (2017), 1233--1252.

\bibitem{kovacs2014b}
{\sc Kovacs, A.~A., Golden, B.~L., Hartl, R.~F., and Parragh, S.~N.}
\newblock Vehicle routing problems in which consistency considerations are
  important: A survey.
\newblock {\em Networks 64}, 3 (2014), 192--213.

\bibitem{kovacs2015a}
{\sc Kovacs, A.~A., Golden, B.~L., Hartl, R.~F., and Parragh, S.~N.}
\newblock The generalized consistent vehicle routing problem.
\newblock {\em Transportation Science 49}, 4 (2015), 796--816.

\bibitem{kovacs2014a}
{\sc Kovacs, A.~A., Parragh, S.~N., and Hartl, R.~F.}
\newblock A template-based adaptive large neighborhood search for the
  consistent vehicle routing problem.
\newblock {\em Networks 63}, 1 (2014), 60--81.

\bibitem{kovacs2015b}
{\sc Kovacs, A.~A., Parragh, S.~N., and Hartl, R.~F.}
\newblock The multi-objective generalized consistent vehicle routing problem.
\newblock {\em European Journal of Operational Research 247}, 2 (2015),
  441--458.

\bibitem{LPS2015}
{\sc Lam, S.~K., Pitrou, A., and Seibert, S.}
\newblock Numba: A llvm-based python jit compiler.
\newblock In {\em Proceedings of the Second Workshop on the LLVM Compiler
  Infrastructure in HPC\/} (New York, NY, USA, 2015), LLVM '15, ACM,
  pp.~7:1--7:6.

\bibitem{lian2016}
{\sc Lian, K., Milburn, A.~B., and Rardin, R.~L.}
\newblock An improved multi-directional local search algorithm for the
  multi-objective consistent vehicle routing problem.
\newblock {\em IIE Transactions 48}, 10 (2016), 975--992.

\bibitem{lim2007}
{\sc Lim, A., and Zhang, X.}
\newblock A two-stage heuristic with ejection pools and generalized ejection
  chains for the vehicle routing problem with time windows.
\newblock {\em INFORMS Journal on Computing 19}, 3 (2007), 443--457.

\bibitem{luo2015}
{\sc Luo, Z., Qin, H., Che, C., and Lim, A.}
\newblock On service consistency in multi-period vehicle routing.
\newblock {\em European Journal of Operational Research 243}, 3 (2015),
  731--744.

\bibitem{LV2011}
{\sc Luxen, D., and Vetter, C.}
\newblock Real-time routing with openstreetmap data.
\newblock In {\em Proceedings of the 19th ACM SIGSPATIAL International
  Conference on Advances in Geographic Information Systems\/} (New York, NY,
  USA, 2011), GIS '11, ACM, pp.~513--516.

\bibitem{marzo2020}
{\sc Marzo, R.~G., and Ribeiro, C.~C.}
\newblock A grasp with path-relinking and restarts heuristic for the
  prize-collecting generalized minimum spanning tree problem.
\newblock {\em International Transactions in Operational Research 27}, 3
  (2020), 1419--1446.

\bibitem{nagata2009}
{\sc Nagata, Y., and Br{\"a}ysy, O.}
\newblock A powerful route minimization heuristic for the vehicle routing
  problem with time windows.
\newblock {\em Operations Research Letters 37}, 5 (2009), 333--338.

\bibitem{O15}
{\sc Oliphant, T.~E.}
\newblock {\em Guide to NumPy}, 2nd~ed.
\newblock CreateSpace Independent Publishing Platform, USA, 2015.

\bibitem{rodriguez2019}
{\sc Rodr{\'\i}guez-Mart{\'\i}n, I., Salazar-Gonz{\'a}lez, J.-J., and Yaman,
  H.}
\newblock The periodic vehicle routing problem with driver consistency.
\newblock {\em European Journal of Operational Research 273}, 2 (2019),
  575--584.

\bibitem{solomon1987}
{\sc Solomon, M.~M.}
\newblock Algorithms for the vehicle routing and scheduling problems with time
  window constraints.
\newblock {\em Operations research 35}, 2 (1987), 254--265.

\bibitem{stavropoulou2019}
{\sc Stavropoulou, F., Repoussis, P.~P., and Tarantilis, C.~D.}
\newblock The vehicle routing problem with profits and consistency constraints.
\newblock {\em European Journal of Operational Research 274}, 1 (2019),
  340--356.

\bibitem{tarantilis2012}
{\sc Tarantilis, C.~D., Stavropoulou, F., and Repoussis, P.~P.}
\newblock A template-based tabu search algorithm for the consistent vehicle
  routing problem.
\newblock {\em Expert Systems with Applications 39}, 4 (2012), 4233--4239.

\bibitem{R07}
{\sc van Rossum, G.}
\newblock Python programming language.
\newblock In {\em Proceedings of the 2007 {USENIX} Annual Technical Conference,
  Santa Clara, CA, USA, June 17-22, 2007\/} (2007).

\bibitem{wong2008}
{\sc Wong, R.~T.}
\newblock Vehicle routing for small package delivery and pickup services.
\newblock In {\em The Vehicle Routing Problem: Latest Advances and New
  Challenges}. Springer, 2008, pp.~475--485.

\bibitem{xu2018}
{\sc Xu, Z., and Cai, Y.}
\newblock Variable neighborhood search for consistent vehicle routing problem.
\newblock {\em Expert Systems with Applications 113\/} (2018), 66--76.

\bibitem{zhen2019}
{\sc Zhen, L., Lv, W., Wang, K., Ma, C., and Xu, Z.}
\newblock Consistent vehicle routing problem with simultaneous distribution and
  collection.
\newblock {\em Journal of the Operational Research Society\/} (2019), 1--18.

\end{thebibliography}

\end{document}